\documentclass[runningheads]{llncs}
\usepackage{graphicx}
\usepackage{multirow}
\usepackage{subfig}
\usepackage{hyperref}

\begin{document}
\title{NeuroPapyri: A Deep Attention Embedding Network for Handwritten Papyri Retrieval}


%
%
\author{Giuseppe De Gregorio\inst{1}\orcidID{0000-0002-8195-4118} \and
Simon Perrin\inst{2}\orcidID{0009-0005-2008-2802} \and Rodrigo C. G. Pena\inst{1}\orcidID{0000-0002-9010-2830} \and
Isabelle Marthot-Santaniello\inst{1}\orcidID{0000-0003-0407-8748} \and
Harold Mouchère\inst{2}\orcidID{0000-0001-6220-7216}
}
\titlerunning{NeuroPapyri: Attention Network for Papyri Retrieval}
\authorrunning{G. De Gregorio et al.}
%
\institute{
University of Basel, Basel, Switzerland
\email{\{giuseppe.degregorio,rodrigo.cerqueiragonzalezpena,i.marthot-santaniello\}@unibas.ch}
\and
Laboratoire des Sciences du Numérique de Nantes (LS2N), Nantes Université, École Centrale Nantes, CNRS, LS2N, UMR 6004, F-44000 Nantes, France
\email{\{simon.perrin,harold.mouchere\}@univ-nantes.fr}}
\maketitle              
\begin{abstract}

The intersection of computer vision and machine learning has emerged as a promising avenue for advancing historical research, facilitating a more profound exploration of our past. However, the application of machine learning approaches in historical palaeography is often met with criticism due to their perceived ``black box'' nature. 
In response to this challenge, we introduce NeuroPapyri, an innovative deep learning-based model specifically designed for the analysis of images containing ancient Greek papyri. 
To address concerns related to transparency and interpretability, the model incorporates an attention mechanism. 
This attention mechanism not only enhances the model's performance but also provides a visual representation of the image regions that significantly contribute to the decision-making process. 
Specifically calibrated for processing images of papyrus documents with lines of handwritten text, the model utilizes individual attention maps to inform the presence or absence of specific characters in the input image. 
This paper presents the NeuroPapyri model, including its architecture and training methodology. 
Results from the evaluation demonstrate NeuroPapyri's efficacy 
in document retrieval, showcasing its potential to advance the analysis of historical manuscripts. 

\end{abstract}

\section{Introduction}\label{sec:introcution}
Greek papyri preserved thanks to the dryness of the Egyptian climate are an unrivalled source for historians and philologists, documenting many aspects of daily life but also preserving otherwise lost pieces of literature. However, papyri are most of the time fragmented, spread across collections around the world and lacking information on the context of their discovery. An essential work done by the papyrologists is to reconstruct fragmented papyri. Sometimes the content helps find joins when we can speculate what the text was (known literary text, recurrent formulaic wordings in documents). However, for some important cases like lost pieces of literature, finding similarities in visual layout and shape of handwriting is essential.  
Paleographers stumbled upon the difficulty of defining what they mean by ``similarity''. There is neither consensus on typologies of scripts nor on styles nor even on letter shapes. Literary papyri are not signed so it is only in exceptional cases that we can know their writers. Thus in this study, we used only elements upon which there is consensus: character identification based on transcriptions and line detection based on lines preserved on the same document. 

The intersection of computer vision and machine learning (ML) presents a promising avenue for historical research, enabling a deeper exploration of our past. Nevertheless, ML approaches are often criticized for their perceived "black box" nature. To bridge the gap between paleographers and computer scientists, we introduce NeuroPapyri, an innovative deep learning network tailored to provide an embedding representation of an image of a papyrus written in ancient Greek. NeuroPapyri incorporates an attention layer with multiple attention heads, each specializing in focusing on distinct features within handwritten images. This attention mechanism aims to enhance transparency by revealing the specific regions of interest that influenced the model's decision-making process. The integration of attention mechanisms facilitates more meaningful exchanges between experts in palaeography and computer scientists, fostering collaborative advancements in the field of historical document analysis.
In the following sections, Section~\ref{sec:soa} provides a brief overview of existing methodologies, Section~\ref{sec:model} delves into the architecture of NeuroPapyri and the training methodology employed, Section~\ref{sec:dataset} presents the datasets used for the experimentation. Section~\ref{sec:results} presents the results obtained through rigorous testing. Section~\ref{sec:ablation} is dedicated to an ablation study of the architecture. Finally, Section~\ref{sec:discussion} and~\ref{sec:conclusion} focus on discussion and conclusion. Additional material, including the GitHub repository, is available at the link \href{https://d-scribes.philhist.unibas.ch/en/case-studies/iliad-208/neuropapyri/}{https://d-scribes.philhist.unibas.ch/en/case-studies/iliad-208/neuropapyri/}

\section{Related Works}\label{sec:soa}
The retrieval of papyrus based on fragment analysis has been a relatively under-explored area within the scientific community. In recent years, a surge in interest was rather on the task of Writer Identification, leading to the proposition of several solutions.
In their work, Christlein et al.~\cite{christlein2022writer} delve into the challenges associated with automatic writer identification and retrieval in Greek papyri. The study places significant emphasis on preprocessing techniques and feature sampling, highlighting the essential role of effective binarization in enhancing identification accuracy. Two unsupervised methods, one employing traditional features and the other leveraging self-supervised deep learning, are evaluated. The research underscores the significance of writer identification in advancing studies in linguistics, history, and palaeography related to historical manuscripts. The approach involves local feature extraction using SIFT descriptors and self-supervised learning with a CNN, with binarization considered a critical step. Evaluation on the GRK-Papyri dataset~\cite{grk2019} demonstrates improved accuracy with appropriate binarization.
Peer et al.~\cite{peer_feature_mixing_hip23} propose a deep-learning-based approach for writer retrieval and identification on papyri fragments. The focus is on identifying fragments associated with specific writers and those corresponding to the same image. The authors introduce a novel neural network architecture that combines a residual backbone with a feature mixing stage to enhance retrieval performance. The methodology is evaluated on two benchmarks, PapyRow~\cite{cilia2021papyrow} and HisFragIR20~\cite{seuret2020icfhr}, achieving competitive results in writer retrieval and identification tasks.
Rather than Writer Identification, Pirrone et al.~\cite{pirrone2019papy} address the challenge of assembling fragmented papyri, offering a solution to assist papyrologists in reconstructing historical documents. The proposed method involves a deep siamese network architecture, named Papy-S-Net, designed for matching papyrus fragments. The model is trained and validated on a dataset of 500 papyrus fragments. The study explores various patch extraction approaches, demonstrating superior performance when trained on patches containing text. The method proves effective in a real-use case, achieving a $79\%$ accuracy in matching fragments.
Pirrone et al.~\cite{pirrone2021self} focus on a self-supervised deep metric learning approach for automatically associating ancient papyrus fragments, streamlining the challenging task of fragment reconstruction. The method, based on Deep Convolutional Siamese-Networks, demonstrates superior performance compared to domain adaptation, achieving a notable top-1 accuracy of $87\%$ in a retrieval task on the HisFragIR20 dataset~\cite{seuret2020icfhr}. Emphasizing minimal human intervention, the study suggests that this approach holds promise for aiding papyrologists in fragment reconstruction without the need for extensive manual annotation or reliance on pre-trained models.

\section{The Model}\label{sec:model}
\begin{figure}[ht]
\centering
\includegraphics[width=0.85\textwidth]{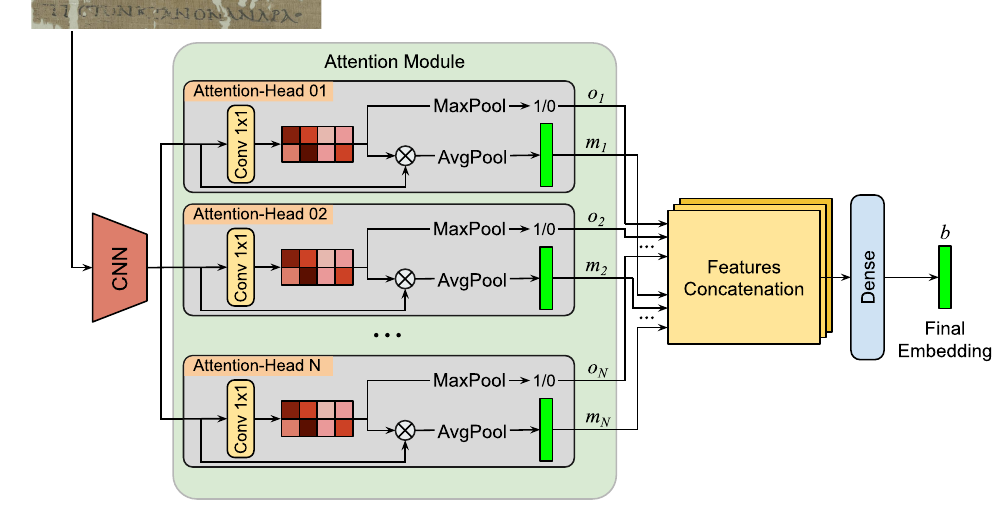}
\caption{Main architecture of the NeuroPapyri model.} \label{fig:mainarchitecture}
\end{figure}

NeuroPapyri is crafted for the task of analyzing images of ancient Greek papyri, aiming to retrieve relevant information and provide an embedding representation. The architectural foundation is established upon an integration of Convolutional Neural Networks (CNN) and a purposefully engineered multi-head attention layer. 
Figure~\ref{fig:mainarchitecture} displays the general architecture of the proposed model.

The model is specifically calibrated for processing images featuring lines of handwritten text. The architecture comprises an initial convolutional phase dedicated to feature extraction, followed by the incorporation of an attention block with multiple heads operating in parallel. As convolutional network, the ResNet-18~\cite{heDeepResidualLearning2015} has been used, as it is recognized for its ability to achieve high performance for a variety of problems in different domains.
At the model's core lies the attention block, a crucial component facilitating simultaneous focus on diverse regions of interest within the image. The attention block features several heads, each attention head functions autonomously, capturing specific information and thereby contributing to the formulation of a comprehensive representation. The underlying idea is to facilitate collaborative functionality among the attention heads, discerning unique strokes, curves, and details essential for accurate comprehension of the manuscript text. Each attention head produces both a feature map $m_i$ and a binary output $o_i$. The computation of the feature map $m_i$ involves a 1x1 convolutional layer designed to condense all features generated by the preceding convolutional phase into a singular attention map. A sigmoid activation function is used on the attention map to allow the values to lie between 0 and 1 and allow the map to pay equivalent levels of attention to different areas of the image, unlike what softmax would do. This map is then multiplicatively applied to the 1x1 convolution input, weighting each pixel of the feature map. The ensuing step involves the amalgamation of distinct feature maps derived from the attention module, which are subsequently subjected to processing by a feed-forward layer, culminating in the accomplishment of the final representation.

The generation of the network's attention maps can be trained with a weakly supervised approach, where the network autonomously refines its attention during training. As the network is trained, it will learn on its own to focus its attention on certain parts of the information to make its decision. However, the model introduces the possibility to enable the guidance of this learning phase through the binary outputs of the attention heads $o_i$. An attention loss is incorporated during the training phase, allowing for deliberate direction of attention map generation. This informative mechanism empowers the model to discern the presence or absence of specific characteristics within different training images. 

\subsection{Retrieval of Original Papyrus}

The main goal of this work is to use the NeuroPapyri network to address the problem of document retrieval within historical papyri written in ancient Greek, recognizing that papyrus images are commonly organized as fragments rather than whole document images. This means that the image of a papyrus may represent a fragment of a larger document, so it raises the question of which original papyrus the fragmentary image corresponds to.

A segmented-based approach that takes into consideration images of the text lines of the document helps to focus on the paleographic features of the manuscript and therefore on the characteristics of handwriting, giving less weight to other features related to elements of the image which may suffer more the effect of different digitizing protocols (such as different brightness levels, different resolutions, and so on).
This aspect is important to take into consideration, as typically the different images of papyrus fragments are preserved in different collections which have not harmonized their protocols and digitization processes.
In this context, therefore, framing the problem as the recovery of a document based on the analysis of lines of handwritten text becomes a relevant consideration.

\begin{figure}[ht]
\centering
\includegraphics[width=0.85\textwidth]{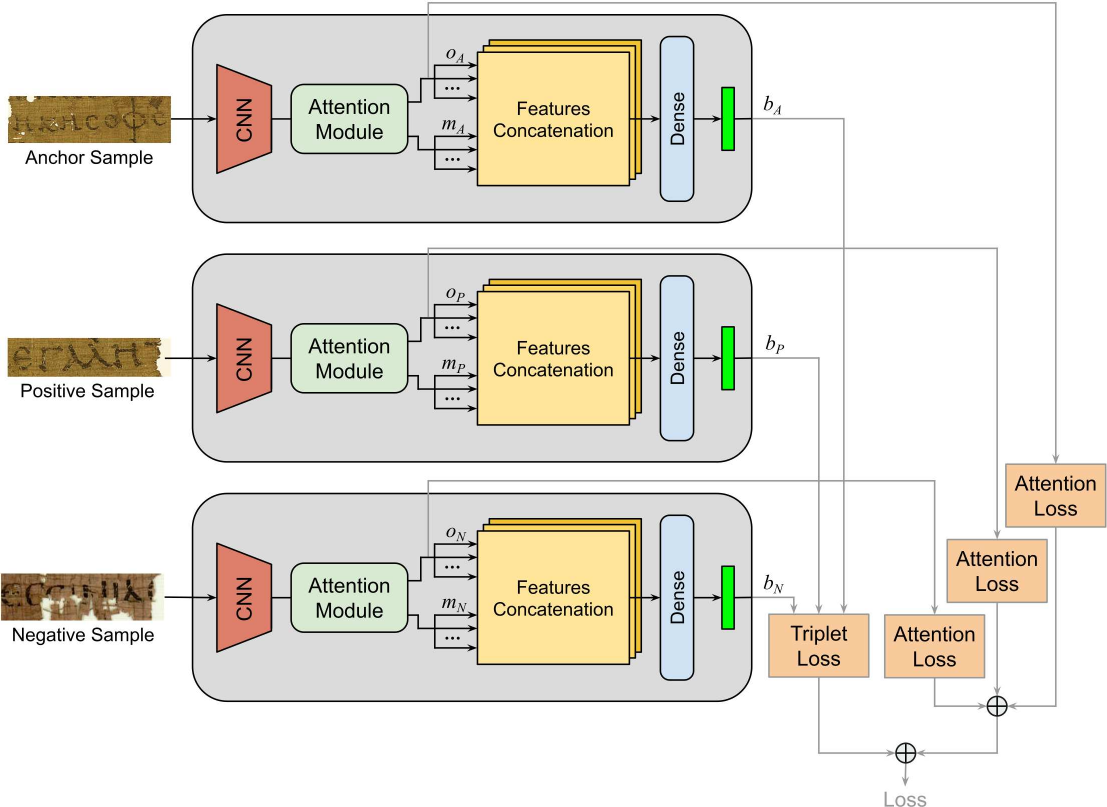}
\caption{NeuroPapyri can be trained using two different losses. \textit{Attention loss} focuses on attention maps while \textit{Triplet loss} aims to train the model to solve the main target problem.} \label{fig:lossarchitecture}
\end{figure}

To address the problem of document retrieval, the proposed model employs triplet learning to derive the final image representation. The architecture is implemented within a Siamese network framework~\cite{chicco2021siamese}, striving to acquire a discerning similarity measure between distinct images. Figure~\ref{fig:lossarchitecture} displays the architecture in the Siamese configuration highlighting the application of the different losses. 
Through the application of a triplet loss, the model is trained to minimize the distance between row representations from the same document, while concurrently maximizing the distance between rows originating from different documents. Consequently, a query image can be associated to a specific document by evaluating its distance from the other images in the dataset.
Notably, the proposed model offers a distinctive feature in its adaptability to two distinct training schemes. In the first scheme, the model autonomously determines attention map configurations, achieved by omitting consideration of attention loss during training. Alternatively, both attention loss ${Loss}_A$ and target loss ${Loss}_T$ can be harnessed concurrently to guide the generation of attention maps. For instance, the model can be directed to focus attention on images containing a specified set of characters, thereby affording a dynamic and tailored approach to training.

In order to afford the model the flexibility to assign distinct levels of importance to different losses, their combination is achieved through a weighted sum, as formulated by the following equation:
\begin{equation}
\label{eq:loss_comin}
Loss = w_1\cdot{Loss}_{A} + w_2\cdot{Loss}_{T}
\end{equation}
Here, $w_1$ and $w_2$ denote the weights assigned to the attention loss and the target loss, respectively. Importantly, these weights are subject to the constraint $w_1+w_2=1$, ensuring that their summation remains constant. The constraint ensures a normalized weighting scheme, facilitating a coherent and interpretable adjustment of loss contributions. This weighted combination of losses allows the model to adjust the influence of each loss during the training process and the model can effectively prioritize either the attention loss or the target loss.

\section{Datasets}\label{sec:dataset}
This section provides insights into two distinct datasets used for experimentation: a Synthetic Dataset and the ICDAR2023 Competition Dataset.

\subsection{Synthetic Dataset}

\begin{figure}
    \centering
    \includegraphics[width=0.6\linewidth]{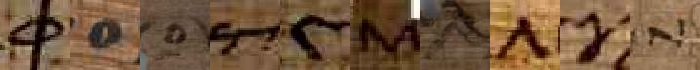}
    \caption{Image from the synthetic dataset. Transcription: $\Phi O O \Pi \Sigma M \Lambda \Lambda \Gamma N $.}
    \label{fig:synthetic_image}
\end{figure}

 A synthetic dataset has been built for the preliminary evaluation stage. To create the synthetic dataset, the AL-PUBv2 dataset~\cite{swindallAlpub2021} has been used. AL-PUBv2 is a dataset containing images of handwritten ancient Greek characters collected by crowdsourcing. 
 The images of this dataset have a very different quality due to preservation state, digitization methods or crowdsourcing operations. 
The synthetic dataset is built by pasting, side by side, 10 images from AL-PUBv2 resulting in $70\times 700$ pixels images. The training set is made of 32768 images with transcription, while the testing set counts 1024 images. First strings of characters are formed randomly, and then images of each of those characters are taken randomly from AL-PUBv2. None of the character images used to build the test set are used in the training set. 
Figure~\ref{fig:synthetic_image} presents an example of a created image. 

\subsection{ICDAR2023 Competition Dataset}

The ``ICDAR 2023 Competition on Detection and Recognition of Greek Letters on Papyri''~\cite{seuret2023icdar} has proposed a dataset containing 194 high-resolution images of papyri. All these papyri bear parts of Homer's \textit{Iliad}. Each papyrus image is annotated by specialists at the character level. A bounding box is drawn around each character associated with its transcription and a quality tag indicating if the character is well preserved, incomplete but unambiguously recognizable or highly damaged. 

Character distribution analysis is a fundamental aspect in the study of textual datasets, offering insights into the prevalence of individual letters and their linguistic significance. The character distribution of the ICDAR2023 dataset, as illustrated in Figure~\ref{fig:character_distribution_icdar}, showcases disparities in the representation of letters. Notably, certain letters exhibit lower occurrences, suggesting a skewed distribution influenced by the infrequency of their use.

While the primary focus of the competition does not coincide with the objective of document retrieval, the thoroughness of the dataset annotations renders it readily adaptable to the scope of our study. Given the scarcity of dedicated datasets tailored to the task of retrieving ancient papyrus documents, the ease of repurposing this dataset persuaded us to use it for our purpose.

\section{Experimentation and Results}\label{sec:results}
To assess the effectiveness of the NeuroPapyri model, a comprehensive series of experiments was undertaken, encompassing diverse datasets comprising papyrus images containing handwritten text in Ancient Greek. 
In this section, before presenting the results of experiments specifically addressing the retrieval of papyrus documents, we assess the network's performance on a preliminary and simpler task: character identification within text lines. This experiment enables us to ascertain whether the network, with its attention mechanism, effectively discerns the presence of specific characters in the image. Additionally, it offers insights into whether the decision is in any way correlated with the textual content of the images.

\subsection{Character Identification}
This evaluation scenario is designed to assess the presence of characters within a line of handwritten text. To tackle this challenge, the model can be trained by integrating a dedicated attention loss for character identification. 
In this setup, the model is settled with attention heads equal to the number of characters targeted for identification. The output $o_i$ of a given attention head corresponds to 1 if the pertinent character is present in the image, or to 0 if the character is absent. Essentially, $o_i$ represents a binary label, facilitating the modelling of this behaviour through the application of a binary cross-entropy cost function. Therefore, for the purposes of the experiment, the model is parameterized with 24 attention heads, each corresponding to a character of the Greek alphabet. 

To evaluate the model, precision, recall and F1-score are used. A detection is considered positive when the network identifies a letter, regardless of how many such letters are in the image or where it puts its attention to identify the character.
This evaluation serves as an investigation into the efficacy of attention loss, offering insights into its influence on the definition of attention maps.

\subsubsection{Synthetic dataset}

Firstly, the model is trained on the synthetic dataset. Binary cross-entropy loss is used as attention loss. The Adam optimizer is used with a learning rate of $5\cdot10^{-5}$ and a batch size of 16 images is used. 

Table~\ref{tab:character_identification_results} shows the results of character identification for the synthetic dataset. The F1-score of $82.14\%$ shows that the model is able to identify characters in a synthetic papyrus line image using attention loss.

\begin{table}[ht]
    \centering
    \begin{tabular}{cccc}
         Dataset & Precision & Recall & F1-score\\\hline\hline
         Synthetic dataset & 78.15 & 86.55 & 82.14 \\
         ICDAR2023 dataset\cite{seuret2023icdar} & 69.4 & 82.37 & 75.33 \\
         \hline\\
    \end{tabular}
    \caption{Precision, Recall and F1-score for the model with synthetic and ICDAR2023 Competition dataset \cite{seuret2023icdar}.}
    \label{tab:character_identification_results}
\end{table}

\subsubsection{ICDAR2023 competition dataset}

\begin{figure}[ht]
\centering
\subfloat[]{\includegraphics[width=0.45\textwidth]{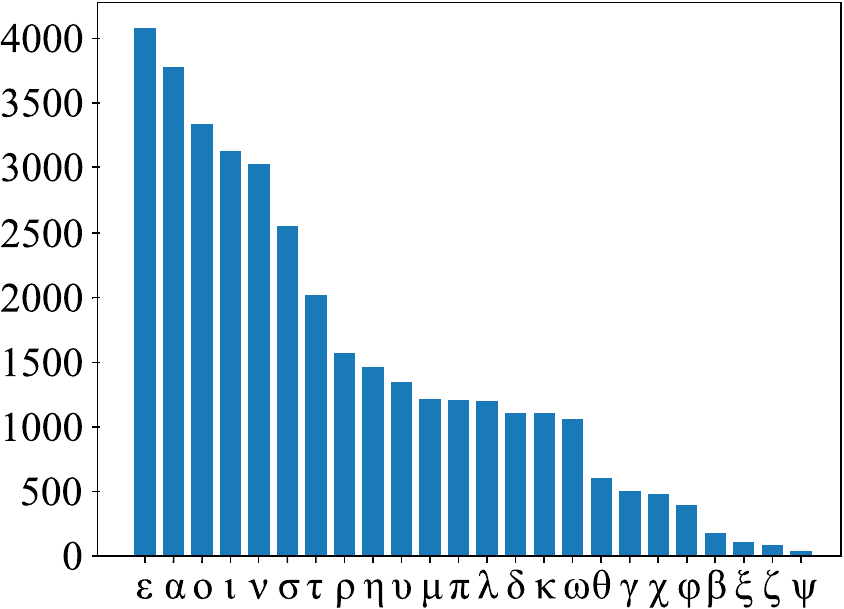}\label{fig:character_distribution_icdar}}
  \hfill
\subfloat[]{\includegraphics[width=0.45\textwidth]{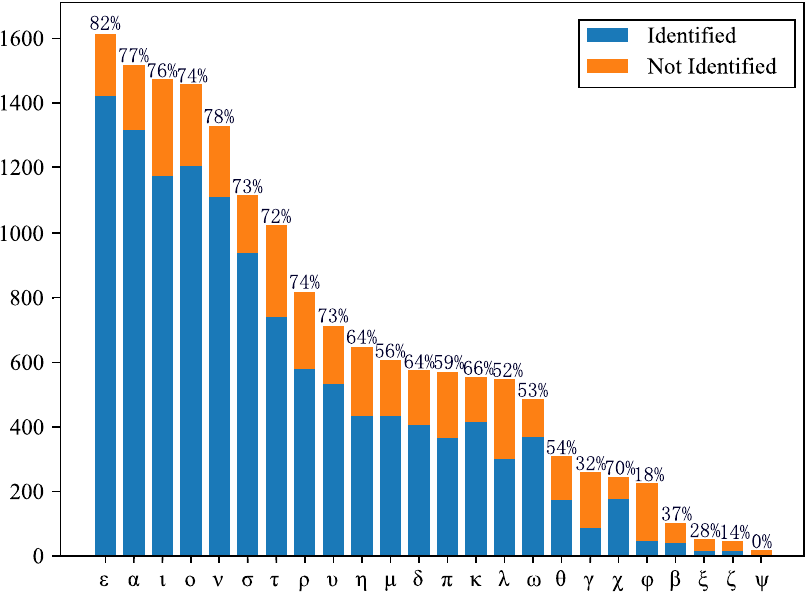}\label{fig:barchart_character_identification}}
\caption{a) Character distribution in the ICDAR2023 competition training set; b) Identification rate for each letter in the ICDAR2023 test set.}
\end{figure}

To evaluate the model on real papyri images, the ICDAR2023 competition dataset is employed. To align the dataset with the problem domain, we meticulously extracted 6,423 text line fragments, each accompanied by its corresponding content transcription from the original images. Labelling was performed in a manner that ensured consistency across lines originating from distinct images of the same document. Subsequently, an $80\%/20\%$ split was applied, designating $80\%$ of the data for training and the remaining $20\%$ for testing. 
Attention loss and Adam optimizer are still used with a learning rate of $1\cdot10^{-4}$ and mini-batches size of 4 images. Greyscale data augmentation is used with a $20\%$ rate. The training lasted 18 epochs.
Results from this experiment are present in Table~\ref{tab:character_identification_results}. Remarkably, the training on real data yields results comparable to those achieved on the synthetic dataset. Figure~\ref{fig:barchart_character_identification} provides a visual representation of the identification rate for each character in the ICDAR2023 test set. We can note that some characters show lower identification rates than others. Characters such as Beta ($B / \beta$), Zeta ($Z /\zeta$), Xi ($\Xi / \xi$), Psi ($\Psi/\psi$) and Phi ($\Phi / \phi$) show poor identification rates but they are also among the fewest represented within the dataset. Lambda ($\Lambda/\lambda$) and Gamma ($\Gamma/\gamma$) also show low identification rates despite their higher representation, potentially influenced by their shapes, which can be visually similar to other characters like Alpha ($A/\alpha$) or Iota ($I/\iota$). In general, it is possible to see how the identification rate is influenced by the level of representation of the character within the dataset. The most present characters show an identification rate which is on average higher than those which are not very frequent in the data set.

\begin{figure}[ht]
\centering
\includegraphics[width=0.7\textwidth]{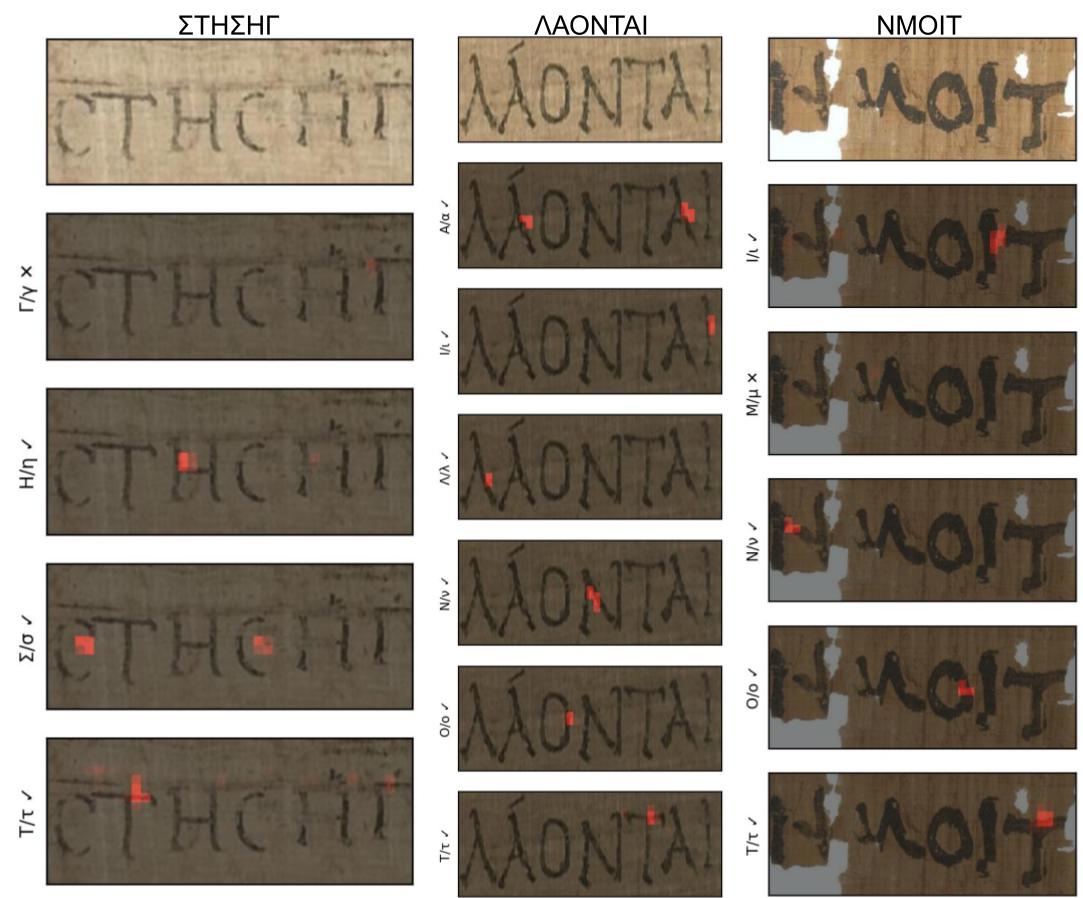}
\caption{Attention maps for some text lines of the ICDAR2023 dataset\cite{seuret2023icdar}.} \label{fig:char_icdar_es}
\end{figure}

Figure~\ref{fig:char_icdar_es} presents some attention maps generated by the network for images from the test set. Notably, the attention map for Eta ($H/\eta$) in the second row first column indicates that the network does focus its attention on the Eta. The attention is only on a small part of the letter, not on the whole letter. The attention maps of Sigma ($\Sigma/\sigma/\varsigma$) and Theta ($\Theta/\theta$) on the first column and of Alpha ($A/\alpha$) on the last column show that the network focuses attention on two characters, suggesting that the network is able to identify multiple characters in an image. Additionally, the attention map dedicated to Mu ($M/\mu$) in the last column shows an instance where the network fails to identify a character, as no attention is allocated to the corresponding region. This Mu was, however, partially damaged.

\subsection{Document Retrieval}
In this section, we present the results of our experiments conducted on the Document Retrieval problem. The dataset employed for this investigation corresponds to the one derived from the ICDAR2023 competition. The used optimization technique is the Adam optimizer, with a learning rate set at $1\cdot10^{-4}$, and a batch size of 32 triplets of images. Cosine distance~\cite{gomaa2013survey} was used to calculate the distance between image embeddings. The training process allowed the network to learn for a maximum of 50 epochs.
While the system effectively generates an ordered list of documents, our emphasis lies in identifying the closest document with meaningful relevance in our specific scenario. Each fragment is associated with a singular document, prompting us to assess the system's performance by considering only the closest document obtained. For this reason, the evaluation metrics reported focus on Top-1 Accuracy and F1@1.

\begin{figure*}[ht]
   \centering
\begin{tabular}{ccc}
  $w_1=0.25$ $w_2=0.75$ & $w_1=0.50$ $w_2=0.50$ & $w_1=0.75$ $w_2=0.25$ \\ 

\includegraphics[width=3.9cm]{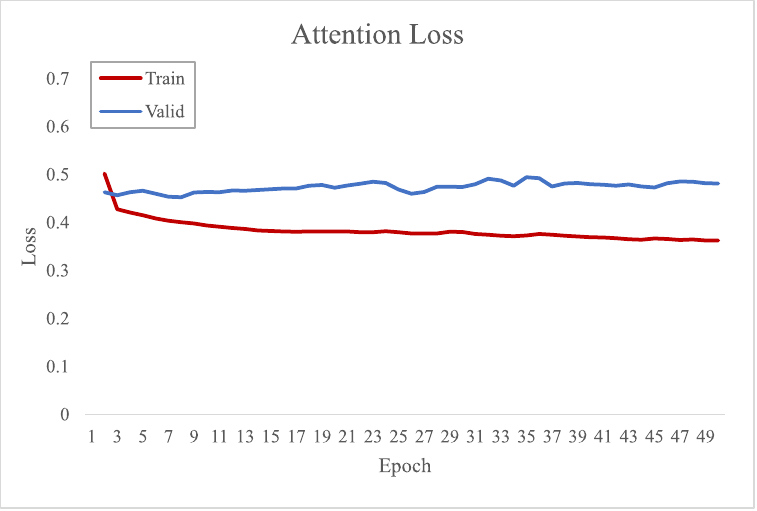}&
\includegraphics[width=3.9cm]{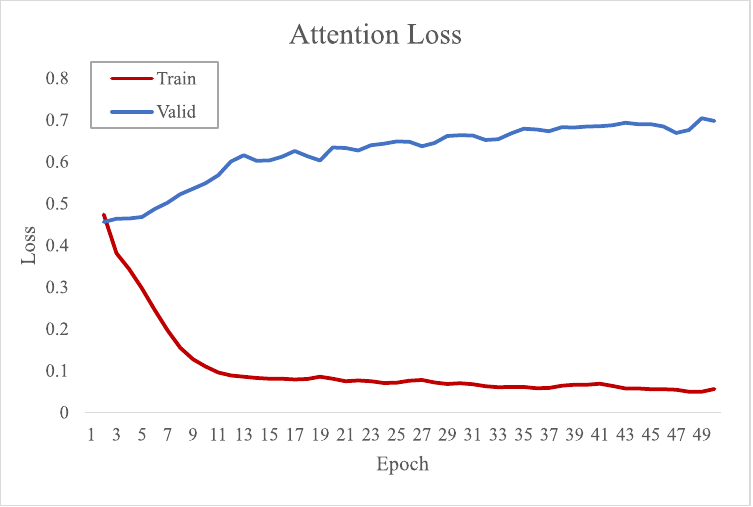}&
\includegraphics[width=3.9cm]{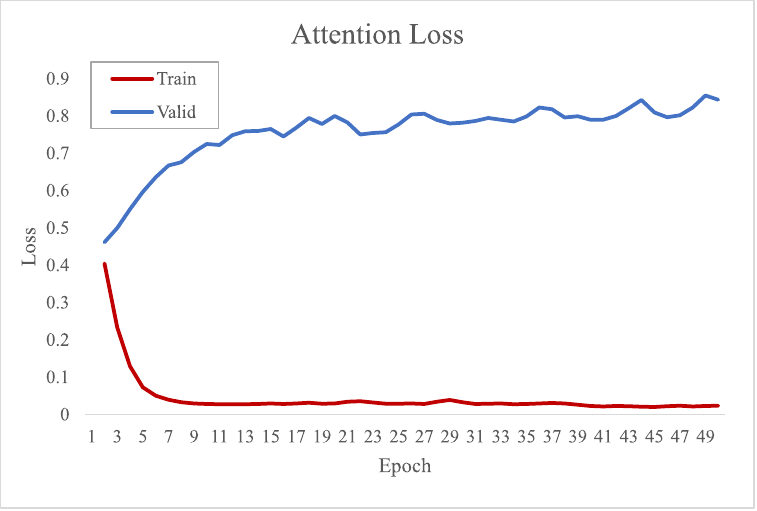}\\

\includegraphics[width=3.9cm]{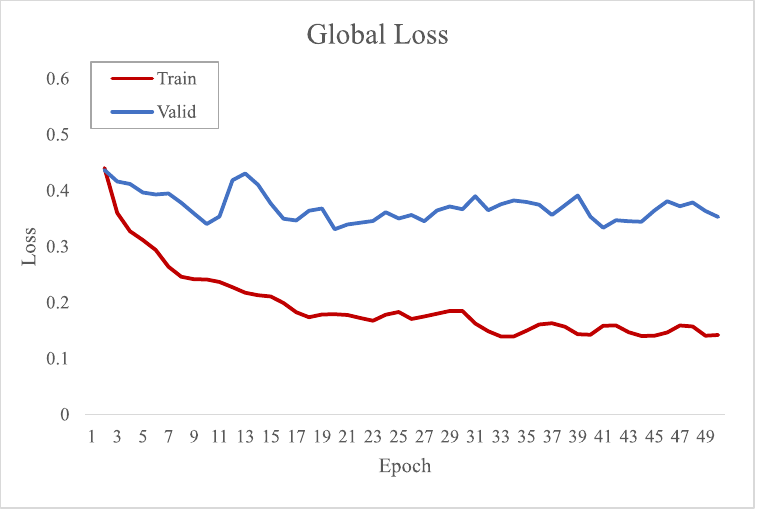}&
\includegraphics[width=3.9cm]{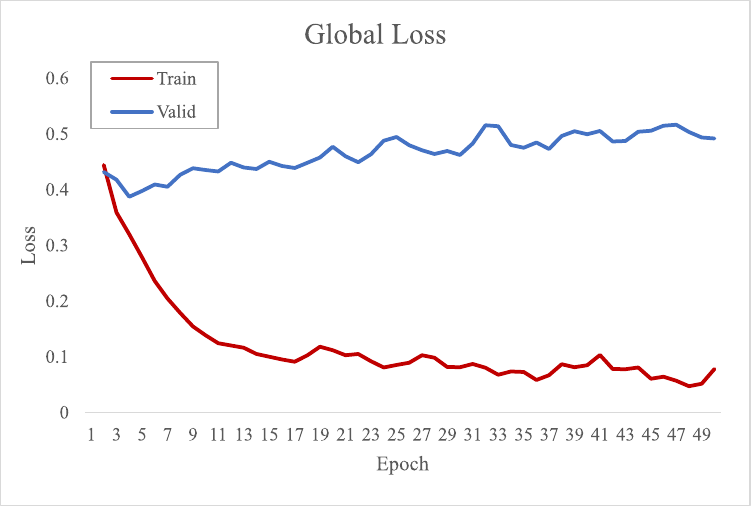}&
\includegraphics[width=3.9cm]{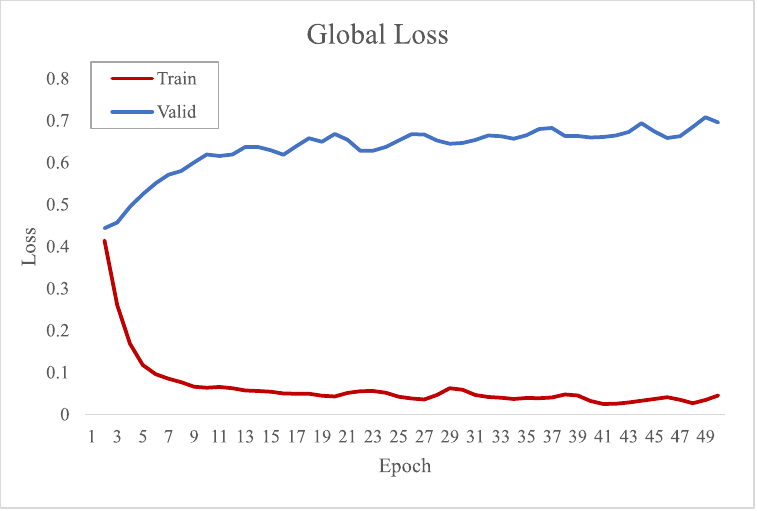}\\
\end{tabular}

    \caption{Evolution of loss during training for different loss weights.}
    \label{fig:losses_attention}
\end{figure*}

As previously indicated, the network can be trained using two distinct methodologies, primarily differing in the incorporation of attention loss and the subsequent modelling of attention map generation. Initially, the network was trained without incorporating attention loss. Subsequently, the model was trained using attention loss, akin to the methodology employed in the character identification experiments. Specifically, each attention head was dedicated to a particular character, with attention maps being conditionally generated based on the presence or absence of the target characters in the image.
Furthermore, the dual loss configuration was explored, as described by equation~\ref{eq:loss_comin}, where the combination of attention loss and target loss is regulated by the weights $w_1$ and $w_2$. Systematic variation of these weights was conducted to assess the influence of each loss on the overall learning trajectory. Experimental findings highlighted a notable divergence in the attention loss when the weight $w_1$ exceeded $0.50$, adversely impacting the system's performance on the validation set. The graphical representation (see Figure~\ref{fig:losses_attention}) illustrates the attention loss trend during training with 24 attention heads as the weights $w_1$ vary. Notably, the loss on the validation set tends to diverge with increasing values of $w_1$.
Table~\ref{tab:loss_weights_res} presents the quantitative results. The performance on the test set demonstrates improvement for lower values of the attention loss weight $(w_1)$, underscoring the importance of attention loss weighting in shaping the system's overall efficacy.

\begin{table}[ht]
   \centering
   \begin{tabular}{cccc}
        & \begin{tabular}[c]{@{}c@{}}$w_1=0.25$\\ $w_2=0.75$\end{tabular} 
        & \begin{tabular}[c]{@{}c@{}}$w_1=0.50$\\ $w_2=0.50$\end{tabular} 
        & \begin{tabular}[c]{@{}c@{}}$w_1=0.75$\\ $w_2=0.25$\end{tabular} \\ 
        \hline\hline

\multicolumn{1}{r|}{\textit{Top1}}   & \textbf{92.82} & 91.47 & 89.71 \\
\multicolumn{1}{r|}{\textit{F1@1}}         & \textbf{89.04} & 88.07 & 84.21 \\
\hline\\
\end{tabular}
\caption{Top-1 Accuracy and F1@1 score of the model as the combination weights of the loss $w_1$ and $w_2$ vary}
\label{tab:loss_weights_res}
\end{table}

\begin{table}[ht]
\begin{tabular}{crccccc}
    & & N-Head=1 & N-Head=3 & N-Head=5 & N-Head=10 & N-Head=24 \\ 
\hline\hline

With  & \textit{Top1} & 95.77 & 94.33 & 93.14 & 92.42  & 91.39  \\
                                   Attention-Loss& \textit{F1@1}       & 92.82 & 90.66 & 89.12 & 86.91  & 88.12  \\
 \hline\hline
Without& \textit{Top1} & \textbf{96.57} & 94.18 & 94.66 & 94.90  & 92.82  \\
                                   Attention-Loss& \textit{F1@1}       & \textbf{94.00} & 93.64 & 91.55 & 91.97 & 89.04  \\ 
\hline
\end{tabular}
\caption{Top-1 Accuracy and F1@1 score of the trained model with and without attention loss and when varying the number of attention heads}
\label{tab:multihead_res}
\end{table}

The determination of the number of attention heads is not hardcoded in the model architecture. Consequently, the model was trained considering diverse numbers of attention heads, ranging up to a maximum of 24, corresponding to the number of letters in the Greek alphabet. Notably, configuring the number of attention heads in the double loss paradigm involves nuanced decisions. A critical consideration in this regard is the selection of distinct attention heads and their association with specific characters. In our experiments, each attention head is intricately linked to the presence of a particular character in the image of the text line.
To delineate the allocation of attention heads, we considered the frequency distribution of letters within the training set. Subsequently, attention heads were selected based on the characters that exhibited the highest frequency of appearance. For instance, in the scenario of a model equipped with three attention heads, the selection process involved considering the three characters that appeared most frequently in the training set ($\varepsilon, \alpha, o$), and so forth.
Table~\ref{tab:multihead_res} provides a comprehensive depiction of the model's performance, showcasing results for both models trained with and without attention loss across varying numbers of attention heads.

\subsubsection{Comparative Investigation.}
To finish this section, we present the results of a comparative investigation conducted between our proposed model and two contemporary methodologies. Specifically, we conducted training procedures on the model introduced by Peer et al.~\cite{peer_feature_mixing_hip23} and the model presented by Pirrone et al.~\cite{pirrone2021self} on the ICDAR2023 dataset, which served as the basis for our evaluation. The tabulated results of this comparative analysis are outlined in Table~\ref{tab:soa_comparison}. Notably, our model exhibited superior performance in contrast to the models under consideration.
The methodology proposed by Peer et al. shares similarities with our approach, as it focuses on the analysis of text lines. Conversely, Pirrone et al.'s model operates on random crops extracted from documents of square dimensions. This methodological divergence raises concerns regarding the consistency of the training data, as it does not ensure the exclusivity of handwritten text within the images, thereby complicating the comparison process.

Nonetheless, it is important to acknowledge the challenges associated with conducting a comprehensive comparative assessment. Limitations arise from the scarcity of bench-marking methodologies and the inherent discrepancies in dataset composition. Furthermore, variations in methodological approaches across different works contribute to the intricacy of bench-marking procedures.

\begin{table}[ht]
\centering

\begin{tabular}{lcc}
Method&Top-1&F1@1\\
\hline
\hline
\textit{Peer et al. \cite{peer_feature_mixing_hip23}} & 71.16 & 70.65\\
\textit{Pirrone et al. \cite{pirrone2021self}} & 88.89 & -\\
\textit{NeuroPapyri} & \textbf{96.57} & \textbf{94.00}\\
\hline
\end{tabular}

\caption{Comparison with State-of-the-Art. Top-1 Accuracy and F1@1 score.}
\label{tab:soa_comparison}
\end{table}

\section{Ablative Study}\label{sec:ablation}
In this section we present an ablative study with the aim to provide valuable insights into the influence of various components of our model on overall performance, allowing us to evaluate whether the different components of the system can contribute positively or negatively to solving the problem. Specifically, we focus on two key characteristics: the adoption of a Siamese architecture and the impact of the proposed attention module.

To establish a baseline for comparison, we trained the model using only the convolutional stage. The outcomes of this experiment serve as a benchmark for evaluating our model's performance and assessing the efficacy of design choices made during the ideation phase. Subsequently, we trained our model based on a Siamese architecture, excluding the proposed attention module. These results are then compared with those obtained by training the complete model, integrating both Siamese architecture and the attention module. Table~\ref{tab:ablation} reports the results of these different model configurations, facilitating conclusions drawn from the ablative study.
Looking at the table, it becomes evident that the adoption of the Siamese architecture alone leads to a significant enhancement in performance, manifesting as a noteworthy 13-point increase in the F1@1 index. Moreover, the introduction of the attention block serves to further refine performance indices, highlighting its significant contribution to the model. 
\begin{table}[ht]
\centering
\begin{tabular}{lcc}
    & Top-1 & F1@1 \\
\hline\hline
\textit{ResNet-18}  &  90.19  &  77.10   \\
\textit{ResNet-18 + Siamese}  &  93.94  &  90.42  \\
\textit{ResNet-18 + Siamese + Attention}  &  \textbf{96.57} & \textbf{94.00} \\
\hline \\
\end{tabular}
\caption{Top-1 Accuracy and F1@1 score for the ablation study}
\label{tab:ablation}
\end{table}

\section{Discussion}\label{sec:discussion}
This section aims to discuss the implications and limitations arising from the findings and methodology presented in this study.

\begin{figure}[ht]
    \centering
    \includegraphics[width=0.68\textwidth]{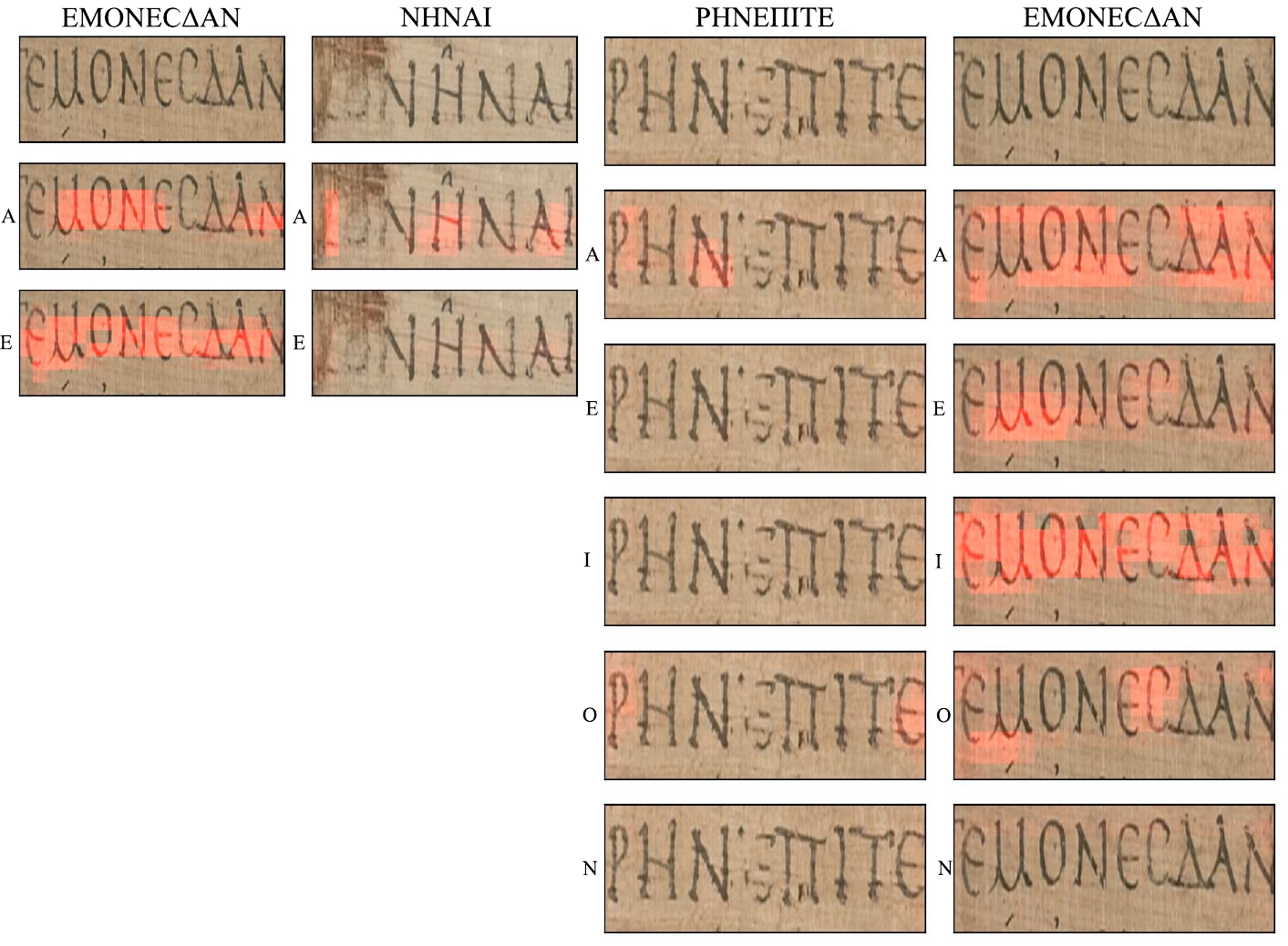}
    \caption{Attention maps for some text lines. The first two images are the results of the system configured with two attention heads (the first for the letter A and the second for the E), while the last two are of a system with five attention heads (for the letters A,E,I,O,N).}
    \label{fig:attentionmaps_retr}
\end{figure}

The incorporation of a multi-head attention mechanism in NeuroPapyri provides insight into an interpretability study of network behaviour, a critical aspect often lacking in machine learning models. Figure~\ref{fig:attentionmaps_retr} shows the attention maps when the network is trained to solve the Document Retrieval problem, where the parts of the image where attention is concentrated are highlighted in red. Observing the attention maps reveals a consistent trend of activation towards the text, indicating the significance of handwriting amidst background elements. Notably, the attention maps exhibit a broader scope of interest beyond solely the target character, often encompassing a segment of writing larger than the character itself.
Furthermore, it can be seen from Table 3 that the best performance is achieved with only one head of attention. This may suggest that, to optimize performance, the network prefers not to focus on individual characters, but to analyze the writing as a whole.

The attention maps provide insights into the regions of interest that influence decision-making. The analysis of those maps can facilitate the collaboration between experts in palaeography and computer science, promoting a deeper understanding of the ancient Greek manuscript writing styles.

While NeuroPapyri demonstrates promising capabilities, certain limitations must be acknowledged. The model's performance heavily relies on the quality and diversity of training data. Specifically, when analyzing the outcomes of the preliminary test of character identification, a notable trend emerges: the model's effectiveness in identifying individual characters correlates with the frequency of appearance of each character within the training set. Characters that occur more frequently in the training data tend to exhibit superior identification performance compared to those with limited representation. This observed pattern in the performance of character identification could potentially influence the distinctive contributions made by each attention head to the broader document retrieval problem. 
Examining the results of the document retrieval, it becomes evident that employing 24 attention heads may not be conducive to optimal performance. Furthermore, the distribution of attention does not fully align with the expectation that importance is given to each individual character. Rather, the network prefers a more general view of the text, trying only to avoid information belonging to the background of the papyrus.

\section{Conclusion}\label{sec:conclusion}
In this work, we have presented a novel deep learning-based model, NeuroPapyri, designed for the analysis of images of ancient Greek papyri. The primary objective of the model is to provide an embedded representation of an image containing handwritten text in ancient Greek through the use of an attention mechanism capable of improving the performance of the model and at the same time providing a visual representation of the areas of the image that contributed most to the decision process.

The model was trained to address the problem of document retrieval. Using a Siamese network framework, the model reached promising results achieving a Top-1 accuracy of $96.57\%$ on the dataset used for the ICDAR 2023 ``Competition on Detection and Recognition of Greek Letters on Papyri''.
The innovative inclusion of an attention block further refined the model's performance, allowing it to learn details and relationships within handwritten text. The attention block generates visible maps that could allow scholars to evaluate which areas of the image contributed most to the decision. Through systematic experimentation, we explored the impact of attention block on system performance, recording an actual improvement in results.
The model's effectiveness in character identification was demonstrated through experiments on two distinct datasets. By employing 24 attention heads, each corresponding to a letter in the Greek alphabet, NeuroPapyri achieved promising results showcasing the model's ability to focus attention on identified letters within the images. 

The development of NeuroPapyri opens avenues for future research directions. Further refinement of attention mechanisms and exploration of advanced architectures may improve the model's robustness and adaptability to diverse historical datasets. The impact of losses on the learning trajectory deserves a more in-depth analysis to understand the best way to combine the different losses and the real contribution each loss makes to the learning process. NeuroPapyri's adaptability can be further tested by evaluating the network on different problems, such as Writer Identification or dating of papyri.

In conclusion, NeuroPapyri is a promising deep-learning framework for analysing handwritten papyri in ancient Greek. By combining cutting-edge technologies in computer vision and machine learning, this model contributes to the evolving landscape of historical document analysis, fostering collaboration between experts in palaeography and computer scientists. The transparency provided by the attention mechanism promotes meaningful exchanges and opens avenues for further advancements in the interdisciplinary field of historical research.

%
%
%
\bibliographystyle{splncs04}
\bibliography{ref}
\end{document}